\documentclass{article}
\usepackage{kr}

\usepackage{balance}
\usepackage{times}
\usepackage{helvet}
\usepackage{courier}
\usepackage{color}
\usepackage{bm}
\usepackage{array}
\usepackage{multirow}
\usepackage{multicol} 
\usepackage{xfrac}
\usepackage{algorithmic}
\usepackage[inoutnumbered,linesnumbered,ruled,vlined]{algorithm2e}
\usepackage{amsfonts}

\usepackage[labelsep=period]{caption}
\usepackage{graphicx}
\usepackage{mathrsfs}
\usepackage{epsfig}
\usepackage{epstopdf}
\usepackage{subfigure}
\usepackage{enumerate}
\usepackage{helvet}
\usepackage{url}
\usepackage{amsmath,amssymb,amsthm}
\usepackage{bbding}
\usepackage{pifont}
\usepackage{rotating}
\usepackage{mathtools}
\usepackage{enumitem}

\newcommand{\Freddy}[1]{{\color{black}#1}}
\begin{document}
%\title{Low-Resource Learning via Ontology-based Semantic Composition}
%\title{Compositional Learning for Complex Concept}
\title{Ontology-guided Semantic Composition for Zero-Shot Learning}
%\title{Ontology-Guided Compositional  Low-Resource Learning}
%\title{Transfer Learning Explanation with Ontologies}
\author{
%Jiaoyan Chen\affmark[1], Freddy L\'ecu\'e\affmark[2], Yuxia Geng\affmark[3], Jeff Z. Pan\affmark[4] \and Huajun Chen\affmark[3] \\
Jiaoyan Chen$^1$ \and Freddy L\'ecu\'e$^{2,3}$ \and Yuxia Geng$^4$ \and Jeff Z. Pan$^{5,6}$ \and Huajun Chen$^4$ \\
\affiliations
$^1$Department of Computer Science, University of Oxford, UK \\
$^2$INRIA, France \\
$^3$CortAIx@Thales, Canada \\
$^4$College of Computer Science, Zhejiang University, China \\
$^5$School of Informatics, The University of Edinburgh, UK\\
$^6$Department of Computer Science, The University of Aberdeen, UK
%\affaddr{\affmark[1]Department of Computer Science, University of Oxford, UK} \\
%\affaddr{\affmark[2]INRIA, France; CortAIx@Thales, Canada} \\
%\affaddr{\affmark[3]College of Computer Science, Zhejiang University, China} \\
%\affaddr{\affmark[4]Department of Computer Science, University of Aberdeen, UK}
%Jiaoyan Chen\\
%Department of Computer Science\\
%University of Oxford, UK
%\And 
%Freddy L\'ecu\'e\\
%INRIA, France\\
%CortAIx@Thales, Canada
%
%\And 
% Yuxia Geng \\
% College of Computer Science\\
% Zhejiang University, China 
%\And
%Jeff Z. Pan\\
%Department of Computer Science\\
%University of Aberdeen, UK
%\AND
%Huajun Chen\\
%College of Computer Science, Zhejiang University, China \\
%Alibaba-Zhejiang University Frontier Technology  Research Center 
% \And
%Ian Horrocks\\
%Department of Computer Science\\
%University of Oxford, UK
}
\maketitle

\begin{abstract}
\begin{quote}
Zero-shot learning (ZSL) is a popular research problem that aims at predicting for those classes that have never appeared in the training stage by utilizing the inter-class relationship with some side information.  
In this study, we propose to model the compositional and expressive semantics of class labels by an OWL (Web Ontology Language) ontology, 
%which is more expressive than the traditional class label text and attribute annotations, 
and further develop a new ZSL framework with ontology embedding.
The effectiveness has been verified by some primary experiments on animal image classification and \Freddy{visual} question answering.
\end{quote}
\end{abstract}

\section{Introduction}

Machine learning often relies on a large data set, especially for training deep models. 
However, training and prediction \Freddy{with limited} data (a.k.a low-resource learning) is common in applications such as 
labeling images with new concepts \cite{nikolaus2019compositional}, learning embeddings of new relations in a knowledge graph \cite{qin2020generative},
extracting long-tail information from text \cite{wu2008information},
\Freddy{and Visual Question Answering (VQA) \cite{DBLP:conf/eccv/TeneyH18} with outside knowledge \cite{okvqa}}. 
In supervised classification, prediction with new classes that have never appeared in the training data is often known as \textit{zero-shot learning} (ZSL) \cite{palatucci2009zero}.
Those new classes are called \textit{unseen classes}, while those classes that have labeled training samples are called \textit{seen classes}.

One popular solution for ZSL is 
building the classifier of an unseen class by combining the classifiers (or learned model parameters) of seen classes with some side information,
%\TODO{[What is `class side information'? Or do we mean `side information' rather than `class side information']}
%Widely used side information 
including \textit{(i)} text (e.g., \cite{qin2020generative} learns the embedding of zero-shot relations via relation text description, \cite{lei2015predicting} predicts images of unseen classes via class text description),
%
% for learning the embedding of zero-shot relations~\cite{qin2020generative} in a knowledge graph,
\textit{(ii)} human annotations (e.g., \cite{lampert2009learning} classifies images of new classes via their annotations of visual characteristics),
and \textit{(iii)} graph data (e.g., \cite{kampffmeyer2019rethinking} aligns image classes with WordNet nodes and predicts for unseen classes by Graph Neural Network (GNN) while \cite{gengexplainable} further extends the idea with explanation).
However, the semantic expressiveness of such side information is often too week to represent complex inter-class relationships, thus limiting the  performance.

Inspired by compositional learning, where learned components are combined by symbolic operations for a new component (e.g., the language model by \cite{socher2012semantic} represents a phrase or a sentence by recursively combing the vectors of its words via its parse tree),
we propose to formally represent the semantics of each class label in ZSL by %the composition of atomic components using an 
OWL 2 (Web Ontology Language) class definitions.
With the ontology we first embed its logic axioms, textual information and paths, 
and encode classes into vectors (i.e., semantic encodings).
%with their semantics kept.
We then follow a mapping-based ZSL paradigm 
which learns a mapping function from the input to the semantic encoding via the training samples, 
and predicts by comparing the input's semantic encoding with the semantic encodings of class labels.
%of class labels.
%\TODO{A brief example of how this works (e.g., a summary of the whale example) would be super useful here for KR people}
For example, given an image of a whale, we map it to a semantic encoding and compare it to those semantic encodings of class labels (Blue Whale, Humpback Whale and Killer Whale) -- the nearest neighbour is adopted as the image's class label. 
%where the mapping from image to label semantic encoding can always be learned even some class label e.g., Killer Whale is not in the training data.

This paper \Freddy{introduces} a new ontology-guided neural-symbolic design pattern for the sample shortage problem in machine learning. 
Specially it develops a framework for using OWL ontology as an expressive side information for improving ZSL.
Although this study is in an early stage, some promising results have been achieved in our evaluation with animal image classification and \Freddy{visual} question answering.
%We evaluate our method on animal image classification,
%with two different but representative cases: \textit{(i)} the unseen class is a sibling of those seen classes, which is often fine-grained and disjoint with others,
%and \textit{(ii)} the unseen is an ancestor class of some of those seen classes, which is often general.  
%The former corresponds to the prediction with an unseen species in taxonomy, while the latter corresponds to the  prediction with a family or genus.
%

\section{Background}

\subsection{OWL Ontology}

%\TODO{Briefly Introduce OWL ontology and OWL 2 EL ($\mathcal{EL}^{++}$ Description Logics) @Jeff}

 In this paper, we adopt OWL 2 EL ($\mathcal{EL}^{++}$ Description Logic) ontologies \cite{BaaBL05}. A signature $\Sigma$, noted 
% ELAGATE , defined by 
$(\mathcal{CN}, \mathcal{RN}, \mathcal{IN})$, consists of $3$ disjoint sets of \textit{(i)} atomic concepts $\mathcal{CN}$, \textit{(ii)} atomic relations $\mathcal{RN}$, and \textit{(iii)} individuals $\mathcal{IN}$. 
% DL Constructs
Given a signature, the top concept $\top$, the bottom concept $\bot$, an atomic concept $A$, an individual $a$, an atomic relation $r$, $\mathcal{EL}^{++}$ concept expressions $C$ and $D$ can be composed with 
% ELAGATE the following 
the following constructs:
\begin{equation}
\top\;|\;\bot\;|\;A\;|\;C\sqcap D\;|\;\exists r.C\;|\;\{a\}\nonumber
\end{equation}

An OWL 2 EL ontology
consists of a TBox and an ABox,
where the TBox is a finite set of   %concept axioms (such as $Mammal\sqsubseteq Animal$) and
%relation axioms (such as domain($eat$)=$Animal$), and the ABox is a finite set of
%assertions declaring instances of concepts and relations. $\mathcal{EL}^{++}$ supports 
General Concept Inclusion axioms (e.g. $C \sqsubseteq D$),
% with $C$ is subsumee and $D$ subsumer)
% ELAGATE \footnote{For the sake of clarity we consider atomic subsumers in GCIs.}) 
%
relation inclusion and composition axioms (e.g., $r \sqsubseteq s$, $r_1\circ \cdots \circ r_n \sqsubseteq s$), etc.
%If C \sqsubsetetq D and D v C, we write C   D. 
% ABox
% An ABox is a set of concept assertion axioms e.g. $a : C$, role assertion axioms e.g., $(a; b) : R$, and individual equality axioms e.g., $a = b$, and individual inequality axioms, e.g., $a \neq b$.
An ABox is a set of concept assertion axioms,
%e.g., $a : C$, 
relation assertion axioms,
%e.g., $(a; b) : r$, 
individual in/equality axioms, etc.
%e.g., $a \neq b$ or $a = b$. 
%A concept definition $A \doteq C$ is an abbreviation  form  of the two GCIs  $A\sqsubseteq C$ and $C\sqsubseteq A$.

\subsection{The Problem of ZSL}
In machine learning classification, a classifier is trained to approximate a target function $f: x \rightarrow y$, where $x$ denotes the input data, $y$ denotes the output i.e., class label.  
ZSL is to classify data whose possible class labels have been omitted from the training data \cite{palatucci2009zero,xian2018zero}.
Formally, we denote \textit{(i)} the samples for training as $\mathcal{D}_{tr} = \{(x, y) | x \in \mathcal{X}_s, y \in \mathcal{Y}_s\}$, where $\mathcal{X}_s$ and $\mathcal{Y}_s$ represent the training data and seen class labels, %(i.e., seen classes) 
respectively;
\textit{(ii)} the samples for testing (prediction) as $\mathcal{D}_{te} = \{(x, y) | x \in \mathcal{X}_u, y \in \mathcal{Y}_u\}$\footnote{The label set of $\mathcal{D}_{te}$ is often set to $\mathcal{Y}_u \cup \mathcal{Y}_s$, a.k.a generalized ZSL. Our method can deal with both ZSL and generalized ZSL.
%but is evaluated with ZSL in the current stage.
},
where $\mathcal{X}_u$ and $\mathcal{Y}_u$ represent the testing data and unseen class labels,  respectively, with $\mathcal{Y}_u \cap \mathcal{Y}_s = \emptyset$;
and \textit{(iii)} the side information as a function $h: y \rightarrow z$, where $z$ represents the semantic encoding of class label $y$.
Our ZSL problem is to predict the class labels of samples in $\mathcal{X}_u$ as correctly as possible.
%Our ZSL problem is to learn the target function $f$ with $\mathcal{D}_{tr}$ and $\left\{ h(y) | y \in \mathcal{Y}_u \cup \mathcal{Y}_s\right\}$, 
%such that the label predictions of samples in $\mathcal{X}_u$ are as correct as possible. 

\section{Methodology}
\begin{figure*}[!t]
\centering
\includegraphics[width=6.7in]{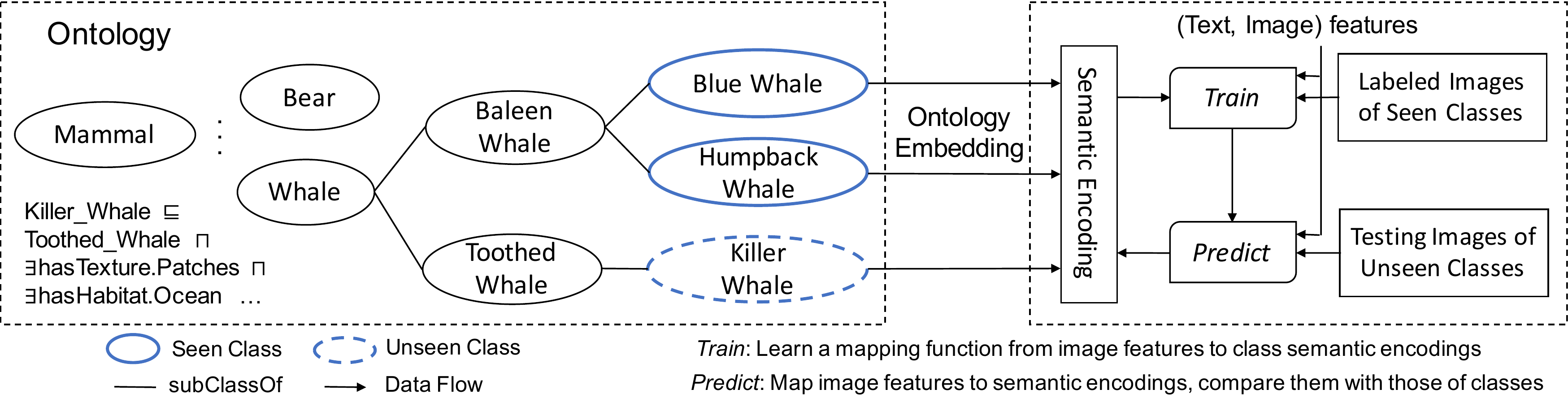}
\caption{The ZSL Framework with Ontology and Ontology Embedding. The framework is illustrated on an image classification task but could be adapted to any machine learning architectures in \textit{Train} and \textit{Predict}. For instance we could envision applying ZSL on a VQA task where joint features of both question text and image are used as the input.}
\label{fig:framework}
\end{figure*}

%The framework of our ZSL method is shown in Figure \ref{fig:framework}.
As shown in Figure \ref{fig:framework}, our ZSL framework includes \textit{(i)} an OWL ontology for representing complex concepts, where each class label corresponds to a concept,
%each of which corresponds to one class label, 
\textit{(ii)} ontology embedding which encodes concepts into vectors with semantics of logic axioms, text and paths, 
and \textit{(iii)} a mapping-based ZSL method.
%
%\Freddy{We mainly describe the semantic encoding and embedding part, as the initial architecture of the learning model does need to be changed.}
\Freddy{We mainly describe the core component of the ZSL framework such as the semantic encoding, 
while the initial architecture for feature learning does need to be changed according to the task.
%as the initial architecture of the learning model does need to be changed.
}

\subsection{Ontology for Semantic Composition}\label{sec:onto}
We build an  OWL 2 EL ontology ~\cite{BaaBL05} to define the corresponding complex concepts for class labels.
%and model their semantic composition mainly using concept hierarchy ($C_1 \sqsubseteq C_2$) conjunction ($C_1 \sqcap C_2$) and existential quantification ($\exists R.C$). 
Take the animal image classification as an example, the
  subsumption can model the taxonomic relations, e.g., $Killer\_Whale \sqsubseteq Toothed\_Whale$.
The existential quantification %together with a set of hierarchical relations 
can define the visual characteristics of animals, e.g., $\exists hasTexture.Patches$ and $\exists hasHabitat.Ocean$.
Concept definitions can then be used to define   composed classes, such as 
%while putting these  two together can compose the semantics of a complex concept, e.g., 
$Killer\_Whale \doteq Toothed\_Whale \sqcap \exists hasTexture.Patches \sqcap \cdots$.
We choose OWL 2 EL due to \textit{(i)} its %expressivity for expressing complex classes,
support of the existential quantification,
\textit{(ii)} polynomial time complexity for entailment reasoning, and \textit{(iii)} and the %limitation of the state-of-the-art
availability of ontology embedding method for OWL 2 EL, rather than %logic operations.
%More details of the last point will be introduced in the following subsection.
%In the future, more expressive KR 
expressive languages  like f-SWRL~\cite{PSTH05}.% can be considered.

In addition to logical axioms,    textual information of concepts in ontologies are useful for ZSL too \cite{qin2020generative,xian2018zero,xian2016latent}. 
It often contains different but complementary semantics as the aforementioned logical  axioms. 
OWL 2 EL ontologies allow the use of rich annotations, including the names, synonyms and descriptions of concepts and relations by e.g., \textit{rdfs:label} and \textit{rdfs:comment}.

%The ontology is specific to 
Given a specific ZSL task,
the supporting ontology  can be either edited by domain experts, or  (more often) semi-automatically created and curated with expert knowledge, third party data and ontologies.
In the above examples: the taxonomyic relations can come from WordNet \cite{miller1995wordnet}; the textual information can be extracted from DBpedia \cite{auer2007dbpedia}; the existential quantification can be created by domain experts or via reusing existing ontologies.

\subsection{Ontology Embedding}
With a given ontology, we implement the label encoding function $h: y \rightarrow z$ by
\textit{(i)} embedding the logical axioms according to their geometric construction \cite{kulmanov2019embeddings}, 
and \textit{(ii)} fine-tuning the word vector model by  textual information and graph paths.
%by fine-tuning the word vector model.

\subsubsection{Embedding of Logical Axioms}
Ontology embedding is to approximate the interpretation of a given ontology by %modeling the logical definitions of concepts and relations
mapping logical axioms into a geometric space,
based on which numeric losses are calculated and vector representations of concepts and relations are learned by optimization.

Specially, a concept $C$ is modeled as a ball, namely a vector denoted as $\nu(C) \in \mathbb{R}^n$, where $n$ is the embedding dimension that can be configured, and a radius denoted as $\gamma(C) \in \mathbb{R}$. A relation $r$ is modeled as a vector $\nu(r) \in \mathbb{R}^n$. 
%\TODO{[Why are the embeddings of a concept and a relation have such similar form?]}
%of the concept and relation.

To extract samples for training, OWL 2 EL axioms are first normalized into four normal forms: $C \sqsubseteq D$, $C \sqsubseteq \exists r.D$, $ \exists r.D \sqsubseteq C$ and $C \sqcap D \sqsubseteq E$.
This can be implemented by OWL ontology reasoners such as jCel \cite{mendez2012jcel}, TrOWL~\cite{TPR2010} and HermiT \cite{glimm2014hermit}.
A loss function which separately deals with the axioms of each normal form is defined, denoted as $\mathcal{L}$.
For simplicity, we present the loss calculation of $C \sqsubseteq D$ and  $C \sqsubseteq \exists r.D$
bellow as they are the most commonly used in defining the complex concept for a ZSL problem:
\begin{equation}\label{eq:CD}
\begin{aligned}
    \mathcal{L}(C \sqsubseteq D) =
    %\mathcal{L}_{C \sqsubseteq D}(c, d) =
    \\ max(0, \parallel\nu (C) - \nu (D)\parallel +  \gamma(C)  - \gamma(D)  - \epsilon)  \\
    + \left| \left\| \nu (C)\right\| - 1  \right| +  \left| \left\| \nu (D)\right\| - 1  \right|,
\end{aligned}
\end{equation}
\begin{equation}\label{eq:CRD}
\begin{aligned}
    \mathcal{L}(C  \sqsubseteq \exists r.D) = max(0, \\
    %\mathcal{L}_{C  \sqsubseteq \exists R.D}(c, d, r) = \\
    %max(0, 
    \parallel\nu (C) + \nu (r) - \nu(D) \parallel +  \gamma(C)  - \gamma(D)  - \epsilon)  \\
    + \left| \left\| \nu (C)\right\| - 1  \right| +  \left| \left\| \nu (D)\right\| - 1  \right|,
\end{aligned}
\end{equation}
%\TODO{[Why don't we give the formula for the other two normal forms?]}
where
%$c$, $d$, $r$ denote the specific concept and relation of an input axiom in training, 
$\left\| \cdot \right\|$ denotes L2-norm, $\left| \cdot \right|$ denotes the absolute value, $max(\cdot, \cdot)$ denotes selecting the maximum value,
$\epsilon$ is a hyper parameter for the max margin loss.
The subsumption $\sqsubseteq$ is modeled by geometric inclusion of two balls (violation will be punished by increasing the loss),
while the existential logic $\exists r.C$ is modeled by translation (i.e., a directed movement of the concept vector).
%\TODO{[Why don't we explain the other EL++ constructs introduced in Sec 2?]}
The concept vectors are also normalized.
Readers are refferred to   \cite{kulmanov2019embeddings} for the loss functions of the   remaining normal forms.% and more details.

In training,  axioms of the normal forms are  extracted from the ontology as samples. 
A stochastic gradient descent algorithm is used to learn the embeddings of concepts and relations by minimizing the overall loss.
The embedding vector of the concept  that corresponds to a ZSL class label $y$ is denoted as $\nu (y)$.

\subsubsection{Embedding of Text and Paths}
We can directly use the word vector of a class label (or the average of vectors of its compositional words), 
using a pre-trained language model  by a general corpus like Wikipedia articles; however, 
  this would ignore the statistical correlation between words and concepts in the   ontology.
Thus we propose to fine-tune the pre-trained model by a   corpus extracted from the ontology which includes the semantics of text and paths, following and extending the idea of OWL2Vec \cite{holter2019embedding}.

To extract the local corpus we first project the original OWL ontology into an RDF (Resource Description Framework) graph
using the method in \cite{agibetov2018supporting}.
%using the method and implementation in \cite{giese2015optique}: 
For example, the concept inclusion axiom $C \sqsubseteq D$ is transformed into $\langle C, \textit{rdfs:subClassOf}, D\rangle$,
while the existential logic axioms $C \sqsubseteq \exists r. D$ and  $\exists r.D \sqsubseteq  C$ are both transformed into $\langle C, r, D\rangle$.
%The approximation loses some semantics of the logic,
%but keeps the
As we only aim at keeping the relative position (statistical correlation) of concepts, relations and words in the ontology, 
such approximations %which fail to keep the logic 
are reasonable. % relative position (statistical correlation) of concepts, relations and words in the ontology.
%(3) ... 
%The axiom $C \sqcap D \sqsubseteq E$ is transformed into $\langle Z, \textit{rdfs:subClassOf},  C\rangle$, $\langle Z, \textit{rdfs:subClassOf},  D\rangle$ and $\langle Z, \textit{rdfs:subClassOf},  E\rangle$, where $Z$ is a fresh new concept representing $C\sqcap D$.
We then apply a random walk algorithm over the graph to extract paths,
and transform them into word sequences as the new corpus by concatenating the concept and relation labels.
%with their order in a path kept.

The corpus from the ontology is finally used to fine-tune a pre-trained language model,
and its resulting vector of a class label $y$ is denoted as $\omega(y)$.
We %A simple way to
merge the embedding of logic axioms and the word vector by concatenating them; i.e., % Namely we have
$h(y) = \left[ \nu(y), \omega(y) \right]$.

\subsection{Mapping-based Training and Prediction}
We adopt a mapping-based ZSL paradigm (see Figure 1) which includes \textit{(i)} the training of a mapping function from the input features (like image features) to the semantic encoding  of seen class labels, 
and \textit{(ii)} the prediction of unseen class labels according to   nearest neighbour.

Instead of directly learning the target function $f:x \rightarrow y$ as in normal supervised learning,
a mapping-based algorithm learns a function from the input to the class label's semantic encoding space, denoted as $g:x \rightarrow z$.
%where $z$ denotes the semantic encoding of labels.
Given the original training samples $\mathcal{D}_{tr}$, it first calculates the semantic encoding of the class labels 
and gets $\mathcal{D}_{tr}^{\prime} = \left\{ (x,z) | z=h(y), (x,y) \in \mathcal{D}_{tr} \right\}$, 
and then uses $\mathcal{D}_{tr}^{\prime}$ to train the function $g$.
Some state-of-the-art mapping-based ZSL algorithms such as DeViSe \cite{frome2013devise} and SAE (Semantic Autoencoder) \cite{kodirov2017semantic} can be adopted. 
Take SAE as an example, briefly it learns a linear encoder from $x$ to $z$ and a decoder from $z$ back to $x$, by minimizing the loss of both the encoder (on $z$) and the decoder (on $x$),
and adopts the encoder as the function $g$.

In prediction, for each testing sample $x_u$ in $\mathcal{X}_u$, the learned function $g$ maps it to a semantic encoding $g(x_u)$,
and $g(x_u)$ then is compared with the semantic encoding of each unseen class label in $\mathcal{Y}_u$.
The unseen class label whose semantic encoding leads to the lowest distance to $g(x_u)$ is determined as the label of $x_u$:
\begin{equation}\label{eq:prediction}
   f(x_u) = \underset{y_u \in \mathcal{Y}_u}{\arg\min} \text{ } dist(h(y_u), g(x_u)), 
\end{equation}
where $dist(\cdot,\cdot)$ denotes a distance function such as L2-norm and cosine similarity.

\section{\Freddy{Experimental Results}}

\Freddy{Preliminary but promising results are reported on two different ZSL tasks to demonstrate the broad potential impact of ontology embeddings.}

\subsection{\Freddy{Settings}}

\vspace{+0.1cm}
\Freddy{\noindent \bf{Animal Image Classification (AIC):}}
\Freddy{We are using the AwA2\footnote{http://cvml.ist.ac.at/AwA2/} \cite{xian2018zero} benchmark, including $50$ mammals} that are partitioned into $40$ seen classes and $10$ unseen classes.
Each class on average has $746$ training / testing images whose $2048$-dimension features (used as the input $x$ in our method) have been extracted by a deep residual neural network.
The corresponding ontology, constructed with the taxonomy from WordNet and expertise of animal visual characteristics (cf. concrete examples in Methodology), includes $1488$ axioms, $180$ classes and $12$ relations.

\begin{figure}[h]
\centering
\includegraphics[scale=0.28]{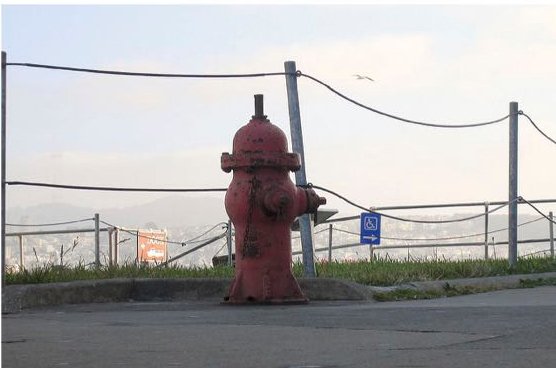}
\vspace{-0.2cm}
\caption{OK-VQA Example. Question: \emph{What vehicle uses this item?} Answer: Firetruck. In our ontology Firetruck is related to FireHydrant by axioms e.g., $Firetruck \sqsubseteq Vehicle \;\sqcap\;\exists\; hasWaterSupply.(ConnectionPoint\;\sqcap FireHydrant)$.}
\label{fig:fire}
\end{figure}

\vspace{+0.1cm}
\Freddy{\noindent \bf{Visual Question Answering (VQA):}}
We are considering a VQA task which is to predict an answer for a given pair of image and question.
The answer is an object (concept) in the image \Freddy{or a Boolean for closed-ended questions}.
Features of the image-question pair are learned from both question text and image by a BERT-based Concept-Vision-Language model (cf. \cite{lu2019vilbert} for the details of the model).
The ZSL framework here is to predict answers that have never appeared in the training data.
%which are composed of image-question pairs and answer labels.
%
\Freddy{
 We adopt the Outside Knowledge-VQA (OK-VQA) dataset \cite{okvqa} which contains $14,031$ images and $14,055$ questions (see example in Figure \ref{fig:fire}). There are $768$ seen classes and $339$ unseen classes.
The ontology is built with ConceptNet knowledge graph \cite{speer2016conceptnet} 
and a core $\mathcal{EL}^{++}$ schema which has been operated on ConceptNet properties that connect answer concepts in OK-VQA questions. 
Specifically $2.3\%$ of ConceptNet properties are involved.}
Figure \ref{fig:fire} presents an example of VQA, where the potential answers of Truck and Firefighter are %equivalent to 
two seen classes in our ZSL setting while Firetruck is %equivalent to 
an unseen class.

\subsection{\Freddy{Evaluation}}

\vspace{+0.1cm}
\Freddy{\noindent \bf{AIC:}}
We calculate the micro accuracy of each unseen class as the ratio of its correctly annotated images,
and then average the micro accuracy of all the unseen classes as the metric \textit{accuracy}.
%, i.e., the metric used in our evaluation.
%which is known as accuracy in the remaining of the paper.
%
Table \ref{res} presents the results of using different semantic encodings: 
Label\_W2V denotes the average word vector of the class label;
Attribute denotes the continuous attribute vector, each of whose slots denotes a real value degree of an attribute annotation on visual characteristics \cite{lampert2009learning};
and EL\_Embed denotes the embedding of ontology logic axioms.
%embedding vector of the class.
Note that the Label\_W2V now adopts \textit{Word2Vec} \cite{mikolov2013distributed} pre-trained by the Wikipedia dump in June 2018,
while the language model fine-tuned by the ontology corpus will be implemented and evaluated in our future work.

\vspace{+0.1cm}
\noindent {\bf{VQA:}} We evaluate the performance of our ZSL framework by comparing the predicted answer with the ground truth answer, 
and calculating the metric \textit{accuracy} as the ratio of the testing image-question pairs whose answers are correct.
Image-question pairs of both seen and unseen answers are tested.
As in AIC, we compare different semantic encodings of the answer label.
The state-of-the-art result comes from \cite{lu2019vilbert} where the word vector of the answer label is adopted \Freddy{(cf. results on VQA in row  Label\_W2V in Table \ref{res}).}
%Label\_W2V are classic word embeddings for question-related features.

%%% \Freddy{
%%% \noindent {\bf{VQA:}}
%
%%%We evaluated the performance of our proposed model using the standard evaluation metric recommended in the VQA challenge \cite{VQA_eval} i.e., $Acc(ans) = \min (1, \sfrac{1}{3}|\{humans\; provided\;ans\}|)$, which measures the accuracy of an answer $ans$. We extended visual- and question-related features extracted from VilBERT \cite{lu2019vilbert} with ConceptNet-related embeddings described in Table \ref{res}. State-of-the-art results are the ones from \cite{lu2019vilbert} where Label\_W2V are classic word embeddings for question-related features.
%%%}

\subsection{Results and Discussion}
In \Freddy{both cases i.e., AIC and VQA (cf. Table \ref{res})}, the embedding of logic axioms outperforms both label word vector and attribute vector, both of which are widely applied and studied semantic encoding solutions in ZSL \cite{xian2018zero}.
Meanwhile, we also find concatenating the embedding to both label word vector and attribute vector significantly improves the accuracy.
For example \Freddy{adding} EL\_Embed improves the accuracy of Label\_W2V by \Freddy{respectively $47.3\%$ and $45.5\%$ in AIC and VQA}.
These observations indicate that the embedding of ontology logic axioms which model the compositional semantics of class labels is an effective and complementary side information for ZSL.
%besides the class label text and attribute annotations.

%
\vspace{+0.1cm}
\Freddy{\noindent {\bf{On AIC-specific Results:}}}
The best ZSL accuracy on \Freddy{AIC case} is now achieved by GNNs \cite{kampffmeyer2019rethinking,wang2018zero}, but they rely on a big graph with around $21$k nodes (concepts) extracted from WordNet.
Such a big graph is often unavailable or hard to be constructed in many real word applications.  
In contrast, our ontology only contains $96$ mammal concepts and uses expressive logic axioms to specify the compositional semantics.
With this small graph, \cite{kampffmeyer2019rethinking} only achieves an accuracy of $6.2\%$ due to sample shortage.
%in training.

%
\vspace{+0.1cm}
\Freddy{\noindent {\bf{On VQA-specific Results:}}
The result with ontology embedding outperforms the state-of-the-art approach, specifically for unseen answers. The semantic encoding has been crucial for encoding semantic relationships among properties and concepts, and derive out-of-knowledge answers i.e., answers which are not given in training data, and only available in the ontology.
}

\begin{table}[h!]
\small{
\centering
\renewcommand{\arraystretch}{1.3}
\begin{tabular}[t]{c|c|c}\hline
Semantic Encoding & AIC & VQA \\ \hline
Label\_W2V & $34.2\%$  &  \Freddy{$30.5\%$}   \\ \hline
Attribute & $47.6\%$ &  \Freddy{$32.1\%$}     \\ \hline
EL\_Embed & $48.3\%$ &  \Freddy{$41.6\%$}    \\ \hline
Label\_W2V + EL\_Embed &$50.4\%$ & \Freddy{$44.4\%$}     \\ \hline
Attributes + EL\_Embed &$56.7\%$ &      \Freddy{$48.1\%$} \\ \hline
Label\_W2V + Attribute &$54.6\%$ &     \Freddy{$34.9\%$} \\ \hline
Label\_W2V + Attribute + EL\_Embed &$58.9\%$ &  \Freddy{$50.1\%$}    \\ \hline
\end{tabular}
\vspace{-0.2cm}
\caption{\small
The Accuracy of SAE with Different Semantic Encodings ($h$). 
$+$ denotes vector concatenation.
}\label{res}
}
\end{table}

\section{Conclusion and Future Work}

In this paper we present a new ZSL framework which utilises an OWL ontology to express the compositional semantics of class labels.
It first learns ontology embeddings that encode the semantics of logical axioms, graph paths and text, 
and then incorporates the embeddings into a mapping-based ZSL paradigm. 
Such a framework provides a new neural symbolic design pattern for dealing with low-resource learning,
via synergistic integration of well expressed formal semantics and neural models.
Some preliminary evaluation results have shown the effectiveness of introducing OWL ontology in ZSL.
In the future, we will \textit{(i)} develop more robust ontology embeddings,
%with text and paths included, 
\textit{(ii)} evaluate the framework with more ZSL algorithms such as the generation based \cite{geng2020generative}, 
and \textit{(iii)} explain the prediction and feature transfer with ontologies \cite{chen2018knowledge}.
%including its graph scale, the logic expressivity~\cite{PRZ2016} and so on. 
%\TODO{Conclusion}

%\TODO{extension of embedding for axioms of OWL EL e.g., for relation domain and range; 

%comparison of two cases: in what kind of contexts the framework is more suitable? 

%Mapping-based ZSL and generation-based ZSL; 

%future work} 

\section*{Acknowledgments}
We want to thank Ian Horrocks from University of Oxford for helpful discussions.
The work is supported by the AIDA project (Alan Turing Institute), the SIRIUS Centre for Scalable Data Access (Research Council of Norway), Samsung Research UK, Siemens AG, and the EPSRC projects AnaLOG (EP/P025943/1), OASIS (EP/S032347/1) and UK FIRES (EP/S019111/1).

%\appendix

%\section{\LaTeX{} and Word Style Files}\label{stylefiles}

%The \LaTeX{} and Word style files are available on the IJCAI--17


\balance
\bibliographystyle{kr}
\bibliography{SemComp}
\end{document}